\documentclass[11pt]{article}
\usepackage{authblk}
\usepackage{eacl2017}
\usepackage{times}
\usepackage{url}
\usepackage{latexsym}
\usepackage{microtype}

\eaclfinalcopy 
\def\eaclpaperid{***} 

\makeatletter
\renewcommand{\paragraph}{%
 \@startsection{paragraph}{4}%
 {\z@}{0.9ex \@plus 1ex \@minus .2ex}{-1em}%
 {\normalfont\normalsize\bfseries}%
}
\makeatother

\title{A Short Review of Ethical Challenges in\\ Clinical Natural Language Processing}

\author[1,2]{\bf Simon \v{S}uster}
\author[1]{\bf St\'{e}phan Tulkens}
\author[1]{\bf Walter Daelemans}
\affil[1]{CLiPS, University of Antwerp}
\affil[2]{Antwerp University Hospital}
\affil[  ]{\tt  \{simon.suster,stephan.tulkens,walter.daelemans\}@uantwerpen.be}

\date{}

\begin{document}
\maketitle
\begin{abstract}
  Clinical NLP has an immense potential in contributing to how clinical practice will be revolutionized by the advent of large scale processing of clinical records. However, this potential has remained largely untapped due to slow progress primarily caused by strict data access policies for researchers. In this paper, we discuss the concern for privacy and the measures it entails. We also suggest sources of less sensitive data. Finally, we draw attention to biases that can compromise the validity of empirical research and lead to socially harmful applications.
\end{abstract}

\section{Introduction}
The use of notes written by healthcare providers in the clinical settings has long been recognized to be a source of valuable information for clinical practice and medical research. Access to large quantities of clinical reports may help in identifying causes of diseases, establishing diagnoses, detecting side effects of beneficial treatments, and monitoring clinical outcomes \cite{Agus2016,Goldacre2014,MurdochAndDetsky2013}. The goal of clinical natural language processing (NLP) is to develop and apply computational methods for linguistic analysis and extraction of knowledge from free text reports \cite{DemnerFushmanEtAl2009,HripcsakEtAl1995,MeystreEtAl2008}. But while the benefits of clinical NLP and data mining have been universally acknowledged, {\it progress in the development of clinical NLP techniques has been slow}. 
Several contributing factors have been identified, most notably difficult access to data, limited collaboration between researchers from different groups, and little sharing of implementations and trained models \cite{ChapmanEtAl2011}. For comparison, in biomedical NLP, where the working data consist of biomedical research literature, these conditions have been present to a much lesser degree, and the progress has been more rapid \cite{CohenAndDemnerFushman2014}.
The main contributing factor to this situation has been the sensitive nature of data, whose processing may in certain situations put patient's privacy at risk.

The ethics discussion is gaining momentum in general NLP \cite{HovyAndSpruit2016}. 
We aim in this paper to gather the ethical challenges that are especially relevant for {\it clinical NLP}, and to stimulate discussion about those in the broader NLP community.
Although enhancing privacy through restricted data access has been the norm, we do not only discuss the right to privacy, but also draw attention to the social impact and biases emanating from clinical notes and their processing. 
The challenges we describe here are in large part not unique to clinical NLP, and are applicable to general data science as well.


\section{Sensitivity of data and privacy}
Because of legal and institutional concerns arising from the sensitivity of clinical data, it is difficult for the NLP community to gain access to relevant data \cite{Barzilay2016,FriedmanEtAl2013}. 
This is especially true for the researchers not connected with a healthcare organization. Corpora with transparent access policies that {\it are} within reach of NLP researchers exist, but are few. An often used corpus is MIMICII(I) \cite{JohnsonEtAl2016,SaeedEtAl2011}. Despite its large size (covering over 58,000 hospital admissions), it is only representative of patients from a particular clinical domain (the intensive care in this case) and geographic location (a single hospital in the United States). Assuming that such a specific sample is representative of a larger population is an example of sampling bias (we discuss further sources of bias in section~\ref{sec:bias}). Increasing the size of a sample without recognizing that this sample is atypical for the general population (e.g.\ not all patients are critical care patients) could also increase sampling bias \cite{KaplanEtAl2014}.\footnote{Sampling bias could also be called selection bias; it is not inherent to the individual documents, but stems from the way these are arranged into a single corpus.} We need more large corpora for various medical specialties, narrative types, as well as languages and geographic areas.

Related to difficult access to raw clinical data is the lack of available {\it annotated} datasets for model training and benchmarking. The reality is that annotation projects do take place, but are typically constrained to a single healthcare organization. Therefore, much of the effort put into annotation is lost afterwards due to impossibility of sharing with the larger research community \cite{ChapmanEtAl2011,FanEtAl2011}. Again, exceptions are either few---e.g.\ THYME \cite{StylerEtAl2014}, a corpus annotated with temporal information---or consist of small datasets resulting from shared tasks like the i2b2 and ShARe/CLEF. 
In addition, stringent access policies hamper reproduction efforts, impede scientific oversight and limit collaboration, not only between institutions but also more broadly between the clinical and NLP communities. 

There are known cases of datasets that had been used in published research (including reproduction) in its full form, like MiPACQ\footnote{The access to the MiPACQ corpus will be re-enabled in the future within the Health NLP Center for distributing linguistic annotations of clinical texts (Guergana Savova, personal communication).}, Blulab, EMC Dutch Clinical Corpus and 2010 i2b2/VA \cite{AlbrightEtAl2013,KimEtAl2015,AfzalEtAl2014,UzunerEtAl2011}, but were later trimmed down or made unavailable, likely due to legal issues. Even if these datasets were still available in full, their small size is still a concern, and the comments above regarding sampling bias certainly apply. For example, a named entity recognizer trained on 2010 i2b2/VA data, which consists of 841 annotated patient records from three different specialty areas, will due to its size only contain a small portion of possible named entities. Similarly, in linking clinical concepts to an ontology, where the number of output classes is larger \cite{PradhanEtAl2013}, the small amount of training data is a major obstacle to deployment of systems suitable for general use.


\subsection{Protecting the individual}
Clinical notes contain detailed information about patient-clinician encounters in which patients confide not only their health complaints, but also their lifestyle choices and possibly stigmatizing conditions. This confidential relationship is legally protected in US by the HIPAA privacy rule in the case of individuals' medical data. 
In EU, the conditions for scientific usage of health data are set out in the General Data Protection Regulation (GDPR). Sanitization of sensitive data categories and individuals' informed consent are in the forefront of those legislative acts and bear immediate consequences for the NLP research.

The GDPR lists general principles relating to processing of personal data, including that processing must be lawful (e.g.\ by means of consent), fair and transparent; it must be done for explicit and legitimate purposes; and the data should be kept limited to what is necessary and as long as necessary.
This is known as data minimization, and it includes sanitization. The scientific usage of health data concerns ``special categories of personal data". 
Their processing is only allowed when the data subject gives explicit consent, or the personal data is made public by the data subject. Scientific usage is defined broadly and includes technological development, fundamental and applied research, as well as privately funded research.

\paragraph{Sanitization} Sanitization techniques are often seen as the minimum requirement for protecting individuals' privacy when collecting data \cite{Berman2002,VelupillaiEtAl2015}. The goal is to apply a procedure that produces a new version of the dataset that looks like the original for the purposes of data analysis, but which maintains the privacy of those in the dataset to a certain degree, depending on the technique. Documents can be sanitized by replacing, removing or otherwise manipulating the sensitive mentions such as names and geographic locations. 
A distinction is normally drawn between anonymization, pseudonymization and de-identification. We refer the reader to Polonetsky et al.\ \shortcite{PolonetskyEtAl2016} for an excellent overview of these procedures. 

Although it is a necessary first step in protecting the privacy of patients, sanitization has been criticized for several reasons. First, it affects the integrity of the data, and as a consequence, their utility \cite{DuquenoyEtAl2008}.
Second, although sanitization in principle promotes data access and sharing, it may often not be sufficient to eliminate the need for consent. This is largely due to the well-known fact that original sensitive data can be re-identified through deductive disclosure \cite{AmblardEtAl2014,DemazancourtEtAl2015,HardtEtAl2016,MalinEtAL2013,Tene2011}.\footnote{Additionaly, it may be due to organizational skepticism about the effectiveness of sanitization techniques, although it has been shown that automated de-identification systems for English perform on par with manual de-identification \cite{DelegerEtAl2013}.} 
Finally, sanitization focuses on protecting the individual, whereas ethical harms are still possible on the group level \cite{ODohertyEtAl2016,TaylorEtAl2017}.
Instead of working towards increasingly restrictive sanitization and access measures, another course of action could be to work towards heightening the perception of scientific work, emphasizing professionalism and existence of punitive measures for illegal actions \cite{FairfieldAndShtein2014,MittelstadtAndFloridi2016}.

\paragraph{Consent} Clinical NLP typically requires a large amount of clinical records describing cases of patients with a particular condition. Although obtaining consent is a necessary first step, obtaining explicit informed consent from each patient can also compromise the research in several ways. 
First, obtaining consent is time consuming by itself, and it results in financial and bureaucratic burdens. It can also be infeasible due to practical reasons such as a patient's death. 
Next, it can introduce bias as those willing to grant consent represent a skewed population \cite{NyrenEtAl2014}.
Finally, it can be difficult to satisfy the informedness criterion: Information about the experiment sometimes can not be communicated in an unambiguous way, or experiments happen at speed that makes enacting informed consent extremely hard \cite{BirdEtAl2016}.

The alternative might be a default opt-in policy with a right to withdraw (opt-out). Here, consent can be presumed either in a broad manner---allowing unspecified future research, subject to ethical restrictions---or a tiered manner---allowing certain areas of research but not others \cite{MittelstadtAndFloridi2016,Terry2012}. Since the information about the intended use is no longer uniquely tied to each research case but is more general, this could facilitate the reuse of datasets by several research teams, without the need to ask for consent each time.
The success of implementing this approach in practice is likely to depend on public trust and awareness about possible risks and opportunities. We also believe that a distinction between academic research and commercial use of clinical data should be implemented, as the public is more willing to allow research than commercial exploitation \cite{Lawrence2016,StaaEtAl2016}.

Yet another possibility is open consent, in which individuals make their data publicly available. Initiatives like Personal Genome Project may have an exemplary role, however, they can only provide limited data and they represent a biased population sample \cite{MittelstadtAndFloridi2016}.

\paragraph{Secure access} Since withholding data from researchers would be a dubious way of ensuring confidentiality \cite{Berman2002}, the research has long been active on secure access and storage of sensitive clinical data, and the balance between the degree of privacy loss and the degree of utility. This is a broad topic that is outside the scope of this article. The interested reader can find the relevant information in Dwork and Pottenger \shortcite{DworkAndPottenger2013}, Malin et al.\ \shortcite{MalinEtAL2013} and Rindfleisch \shortcite{Rindfleisch1997}.

\paragraph{} Promotion of knowledge and application of best-of-class approaches to health data  is seen as one of the ethical duties of researchers \cite{DuquenoyEtAl2008,Lawrence2016}. But for this to be put in practice, ways need to be guaranteed (e.g.\ with government help) to provide researchers with access to the relevant data.  Researchers can also go to the data rather than have the data sent to them.
It is an open question though whether medical institutions---especially those with less developed research departments---can provide the infrastructure (e.g.\ enough CPU and GPU power) needed in statistical NLP. Also, granting access to one healthcare organization at a time does not satisfy interoperability (cross-organizational data sharing and research), which can reduce bias by allowing for more complete input data. Interoperability is crucial for epidemiology and rare disease research, where data from one institution can not yield sufficient statistical power \cite{KaplanEtAl2014}. 

\paragraph{Are there less sensitive data?}
One criterion which may have influence on data accessibility is whether the data is about living subjects or not. 
The HIPAA privacy rule under certain conditions allows disclosure of personal health information of {\bf deceased persons}, without the need to seek IRB agreement and without the need for sanitization \cite{HuserAndCimino2014}.
It is not entirely clear though how often this possibility has been used in clinical NLP research or broader.

Next, the work on {\bf surrogate data} has recently seen a surge in activity. Increasingly more health-related texts are produced in social media \cite{AbbasiEtAl2014}, and  patient-generated data are available online. Admittedly, these may not resemble the clinical discourse, yet they bear to the same individuals whose health is documented in the clinical reports. Indeed, linking individuals' health information from online resources to their health records to improve documentation is an active line of research \cite{PadrezEtAl2015}. Although it is generally easier to obtain access to social media data, the use of social media still requires similar ethical considerations as in the clinical domain. See for example the influential study on emotional contagion in Facebook posts by Kramer et al.\ \shortcite{KramerEtAl2014}, which has been criticized for not properly gaining prior consent from the users who were involved in the study \cite{Schroeder2014}.

Another way of reducing sensitivity of data and improving chances for IRB approval is to work on {\bf derived data}. Data that can not be used to reconstruct the original text (and when sanitized, can not directly re-identify the individual) include text fragments, various statistics and trained models. Working on randomized subsets of clinical notes may also improve the chances of obtaining the data.
When we only have access to trained models from disparate sources, we can refine them through ensembling and creation of silver standard corpora, cf.\ Rebholz-Schuhmann et al.\ \shortcite{RebholzSchuhmannEtAl2011}. 


Finally, clinical NLP is also possible on {\bf veterinary texts}. Records of companion animals are perhaps less likely to involve legal issues, while still amounting to a large pool of data. As an example, around 40M clinical documents from different veterinary clinics in UK and Australia are stored centrally in the VetCompass repository.
First NLP steps in this direction were described in the invited talk at the Clinical NLP 2016 workshop \cite{Baldwin2016}.

\section{Social impact and biases}\label{sec:bias}
Unlocking knowledge from free text in the health domain has a tremendous societal value. However, discrimination can occur when individuals or groups receive unfair treatment as a result of automated processing, which might be a result of biases in the data that were used to train models. The question is therefore what the most important biases are and how to overcome them, not only out of ethical but also legal responsibility. Related to the question of bias is so-called {\it algorithm transparency} \cite{Goodman2016,KamarinouEtAl2016}, as this right to explanation requires that influences of bias in training data are charted. In addition to sampling bias, which we introduced in section 2, we discuss in this section further sources of bias. Unlike sampling bias, which is a corpus-level bias, these biases here are already present in documents, and therefore hard to account for by introducing larger corpora.



\paragraph{Data quality} Texts produced in the clinical settings do not always tell a complete or accurate patient story (e.g.\ due to time constraints or due to patient treatment in different hospitals), yet important decisions can be based on them.\footnote{A way to increase data completeness and reduce selection bias is the use of nationwide patient registries, as  known for example in Scandinavian countries \cite{SchmidtEtAl2015}.} As language is situated, a lot of information may be implicit, such as the circumstances in which treatment decisions are made \cite{HershEtAl2013}.
If we fail to detect a medical concept during automated processing, this can not necessarily be a sign of negative evidence.\footnote{We can take timing-related ``censoring" effects as an example. In event detection, events prior to the start of an observation may be missed or are uncertain, which means that the first appearance of a diagnosis in the clinical record may not coincide with the occurrence of the disease. Similarly, key events after the end of the observation may be missing (e.g.\ death, when it occurred in another institution).} 
Work on identifying and imputing missing values holds promise for reducing incompleteness, see Lipton et al.\ \shortcite{LiptonEtAl2016} for an example in sequential modeling applied to diagnosis classification.

\paragraph{Reporting bias}
Clinical texts may include bias coming from both patient's and clinician's reporting. Clinicians apply their subjective judgments to what is important during the encounter with patients. In other words, there is separation between, on the one side, what is observed by the clinician and communicated by the patient, and on the other, what is noted down. Cases of more serious illness may be more accurately documented as a result of clinician's bias (increased attention) and patient's recall bias. On the other hand, the cases of stigmatized diseases may include suppressed information. In the case of traffic injuries, documentation may even be distorted to avoid legal consequences \cite{IndrayanURL}. 

We need to be aware that clinical notes may reflect health disparities. These can originate from prejudices held by healthcare practitioners which may impact patients' perceptions; they can also originate from communication difficulties in the case of ethnic differences \cite{ZestcottEtAl2016}. Finally, societal norms can play a role. Brady et al.\ \shortcite{BradyEtAl2016} find that obesity is often not documented equally well for both sexes in weight-addressing clinics. Young males are less likely to be recognized as obese, possibly due to societal norms seeing them as ``stocky" as opposed to obese. Unless we are aware of such bias, we may draw premature conclusions about the impact of our results.

It is clear that during processing of clinical texts, we should strive to avoid reinforcing the biases. It is difficult to give a solution on how to actually reduce the reporting bias after the fact.
One possibility might be to model it. If we see clinical reports as noisy annotations for the patient story in which information is left-out or altered, we could try to decouple the bias from the reports. 
 Inspiration could be drawn, for example, from the work on decoupling reporting bias from annotations in visual concept recognition \cite{MisraEtAl2015}. 


\paragraph{Observational bias} Although variance in health outcome is affected by social, environmental and behavioral factors, these are rarely noted in clinical reports \cite{KaplanEtAl2014}.
The bias of missing explanatory factors because they can not be identified within the given experimental setting is also known as {\it the streetlight effect}. In certain cases, we could obtain important prior knowledge  (e.g.\ demographic characteristics) from data other than clinical notes.


\paragraph{Dual use} We have already mentioned linking personal health information from online texts to clinical records as a motivation for exploring surrogate data sources. However, this and many other applications also have potential to be applied in both beneficial and harmful ways. It is easy to imagine how sensitive information from clinical notes can be revealed about an individual who is present in social media with a known identity. More general examples of dual use are when the NLP tools are used to analyze clinical notes with a goal of determining individuals' insurability and employability. 

\section{Conclusion}
In this paper, we reviewed some challenges that we believe are central to the work in clinical NLP. 
Difficult access to data due to privacy concerns has been an obstacle to progress in the field. 
We have discussed how the protection of privacy through sanitization measures and the requirement for informed consent may affect the work in this domain. Perhaps, it is time to rethink the right to privacy in health in the light of recent work in ethics of big data, especially its uneasy relationship to the {\it right to science}, i.e.\ being able to benefit from science and participate in it \cite{Tasioulas2016,Verbeek2014}. We also touched upon possible sources of bias that can have an effect on the application of NLP in the health domain, and which can ultimately lead to unfair or harmful treatment.

\section*{Acknowledgments}
We would like to thank Madhumita and the anonymous reviewers for useful comments.
Part of this research was carried out in the framework of the Accumulate IWT SBO project, funded by the government agency for Innovation by Science and Technology (IWT). 
\bibliography{eacl2017}
\bibliographystyle{eacl2017}

\end{document}